\pdfoutput=1
\documentclass[preprint,12pt,authoryear]{elsarticle}

\usepackage{amssymb}
\usepackage{amsmath}
\usepackage{lineno}
\usepackage{algorithm}
\usepackage{algorithmic}
\usepackage{multirow}
\usepackage[margin=2.5cm]{geometry}
\usepackage[colorlinks,linkcolor=blue]{hyperref}
\usepackage{url}
\usepackage{tabularx}
\usepackage{diagbox}
\usepackage{setspace}
\usepackage{tcolorbox}
\usepackage{caption}
\usepackage{xcolor}
\usepackage{color}
\usepackage{booktabs}
\usepackage{colortbl}
\usepackage{wrapfig}
\usepackage{tablefootnote}
\usepackage{threeparttable}
\usepackage{caption}
\definecolor{cyan}{cmyk}{.3,0,0,0}
\definecolor{mygray}{rgb}{.95,.95,.95}
\newcommand{\MODELNAME}{LHRS-Bot-Nova}
\usepackage{pifont}
\newcommand{\VarSty}[1]{\textnormal{\ttfamily\color{cyan!90!black}#1}\unskip}

\begin{document}

\begin{frontmatter}

\title{\textbf{\MODELNAME: Improved Multimodal Large Language Model for Remote Sensing Vision-Language Interpretation}}
%% Article title


\author[a,b,c,d]{Zhenshi Li}
\author[a,b,c]{Dilxat Muhtar}
\author[a,b,c]{Feng Gu}
\author[a,b,c]{Xueliang Zhang\corref{cor1}}
\author[a,b,c]{Pengfeng Xiao}
\author[f]{Guangjun He}
\author[d,e]{Xiaoxiang Zhu\corref{cor1}}

\affiliation[a]{organization={Jiangsu Provincial Key Laboratory of Geographic Information Science and Technology},
            city={Nanjing},
            postcode={210023},
            country={China}}

\affiliation[b]{organization={Key Laboratory for Land Satellite Remote Sensing Applications of Ministry of Natural Resources},
            city={Nanjing},
            postcode={210023},
            country={China}}

\affiliation[c]{organization={School of Geography and Ocean Science, Nanjing University},
            city={Nanjing},
            postcode={210023},
            country={China}}

\affiliation[d]{organization={Department of Aerospace and Geodesy, Data Science in Earth Observation, Technical University of Munich},
            city={Munich},
            postcode={80333},
            state={Bavaria},
            country={Germany}}

\affiliation[e]{organization={Munich Center for Machine Learning},
            city={Munich},
            postcode={80333},
            state={Bavaria},
            country={Germany}}
            
\affiliation[f]{organization={State Key Laboratory of Space-Ground Integrated Information Technology},
            city={Beijing},
            postcode={100095},
            country={China}}
\cortext[cor1]{Correspondence: \href{zxl@nju.edu.cn}{zxl@nju.edu.cn} (X. Zhang), \href{xiaoxiang.zhu@tum.de}{xiaoxiang.zhu@tum.de} (X.X. Zhu)}

%% Abstract
\begin{abstract}
Automatically and rapidly understanding Earth's surface is fundamental to our grasp of the living environment and informed decision-making. 
This underscores the need for a unified system with comprehensive capabilities in analyzing Earth's surface to address a wide range of human needs.
The emergence of multimodal large language models (MLLMs) has great potential in boosting the efficiency and convenience of intelligent Earth observation. These models can engage in human-like conversations, serve as unified platforms for understanding images, follow diverse instructions, and provide insightful feedbacks. 
In this study, we introduce \MODELNAME, an MLLM specialized in understanding remote sensing (RS) images, designed to expertly perform a wide range of RS understanding tasks aligned with human instructions.
\MODELNAME~features an enhanced vision encoder and a novel bridge layer, enabling efficient visual compression and better language-vision alignment. 
To further enhance RS-oriented vision-language alignment, we propose a large-scale RS image-caption dataset, generated through feature-guided image recaptioning. Additionally, we introduce an instruction dataset specifically designed to improve spatial recognition abilities.
Extensive experiments demonstrate superior performance of \MODELNAME~across various RS image understanding tasks. 
We also evaluate different MLLM performances in complex RS perception and instruction following using a complicated multi-choice question evaluation benchmark, providing a reliable guide for future model selection and improvement.
Data, code, and models will be available at \url{https://github.com/NJU-LHRS/LHRS-Bot}.

\end{abstract}



% \begin{highlights}
% \item 
% \item 
% \item 
% \end{highlights}

%% Keywords
\begin{keyword}



Remote sensing \sep Earth observation \sep Multimodal large language model \sep Vision-language dataset
\end{keyword}

\end{frontmatter}

% \linenumbers

\section{Introduction}
\label{sec:intro}

% Intuitively interpreting Earth observation imagery from remote sensing (RS), 
Interpreting remote sensing (RS) imagery and understanding multi-level features, object relationships, and their dynamic trends, play a significant role in various applications, such as urban sustainable development~\citep{zhang2022direct,wu2023improved,sun2020dramatic}, early-warning systems~\citep{ravuri2021skilful,reichstein2024early,ravuri2021skilful,xu2024large}, and earth surface processes~\citep{qian2024simultaneous,zhu2022remote}. 
Artificial intelligence (AI) has revolutionized RS data analysis~\citep{zhao2024artificial, ma2019deep, reichstein2019deep, zhang2022artificial, zhu2017deep}, and recent advancements in visual foundation models have further improved the efficiency and quality of interpretation of Earth's surface using RS data~\citep{xiong2024neural,guo2024skysense,hong2024spectralgpt}.
However, a major drawback of visual foundation models is their need for tailored designs for specific downstream tasks, leading to fixed functions and limited generalization capabilities. Additionally, they lack the ability to interact with humans, making it difficult to fully address diverse human needs~\citep{zhou2024towards}.

Language, as the primary medium of human communication, plays a fundamental role in facilitating interaction with machines. 
Large language models (LLMs), such as ChatGPT~\citep{ChatGPT}, have demonstrated remarkable conversational abilities, step-by-step reasoning skills, and the capacity to serve as general-purpose task solvers~\citep{touvron2023llama,brown2020language,zhao2023survey}. 
Taking a further step toward human-level AI, multimodal large language models (MLLMs) enhance LLMs with visual perception, enabling them to see and understand the world~\citep{fei2024multimodal,yin2023survey,achiam2023gpt}. 
These models have demonstrated scalability and generalizability as general-purpose assistants~\citep{li2024multimodal}, and have already shown their strong capabilities in understanding RS data for real-world application~\citep{zhang2024good, tan2023promises}.
Developing specialized MLLMs for interpreting RS images offers several advantages:
\textbf{1) unified modeling}: MLLMs provide a versatile framework for handling a wide range of visual tasks across different granularities;
\textbf{2) human-computer interaction}: MLLMs can interpret human intent and incorporate auxiliary information through conversational interactions;
\textbf{3) reasoning}: advanced reasoning capabilities, such as chain-of-thought methods~\citep{wei2022cot, mitra2024compositionalcot}, enable MLLMs to understand complex relations and deal with complex scenarios; and
\textbf{4) enhanced multimodal task potential}: pretrained on extensive and diverse datasets, MLLMs establish a robust baseline for addressing complex multimodal problems.

The potential of specialized MLLMs has been widely acknowledged by the research community~\citep{moor2023foundation, zhou2024towards, li2024vision}, prompting several early efforts to develop large-scale vision-language datasets and RS-specific MLLMs~\citep{muhtar2024lhrs, kuckreja2024geochat, zhang2024earthgpt}.
However, we have identified three main drawbacks in current studies.
\textbf{1) Lack of high-quality, large-scale image-caption dataset:}
High-quality vision-language pretraining datasets are crucial for developing robust multimodal models~\citep{nguyen2022quality, fang2022data, gadre2024datacomp, xu2023demystifying}. Several extensive RS image-caption datasets have been developed to enhance vision-language training~\citep{wang2024skyscript, zhang2023rs5m, muhtar2024lhrs}. However, these datasets often suffer from noisy and uninformative captions, limited semantic richness, poor sentence diversity, and an overfocus on salient objects, which undermine effective modality alignment for RS MLLMs.
\textbf{2) Weaknesses in spatial recognition and hallucination tendencies:} 
We discover that current RS MLLMs exhibit low accuracy in spatial positioning and frequently produce hallucinated responses~\citep{bai2024hallucination} when confronted with questions beyond their capabilities.
\textbf{3) Challenges in holistically evaluating MLLMs:}
The ability of MLLMs to solve various visual tasks serves them as versatile multi-taskers.
Although they often excel in metrics for common tasks like classification, visual question answering, and visual grounding, these metrics fall short of fully reflecting the all-round capabilities of MLLMs—particularly in recognizing complex scenes, objects, attributes, spatial relationships, and, most importantly, following human instructions.
% LHRS-Bench~\citep{muhtar2024lhrs}, which is composed of multiple-choice questions (MCQs), was proposed to thoroughly access RS MLLMs. 
% However, the use of substring matching accuracy as an evaluation metric can introduce significant uncertainty in determining the correctness of an answer, leading to biased accuracy measurements.

In this study, we mitigate the above issues and propose \MODELNAME, an improved RS-specialized MLLM for holistic interpreting RS images with human instructions~\citep{muhtar2024lhrs}. 
\MODELNAME~can respond to user instructions and achieve various RS interpretation tasks with state-of-the-art performance.
To enhance RS-oriented vision-language alignment, we construct a large-scale RS image-caption dataset, LHRS-Align-Recap, by prompting an off-the-shelf multimodal captioner with RS images paired with their OpenStreetMap (OSM) features.
Compared to captions generated solely by a language model using textual information~\citep{muhtar2024lhrs}, utilizing a vision-capable captioner results in richer captions with significantly enhanced language richness and improved alignment between images and captions (Table \ref{tab:dataset stat}). It also provides more detailed descriptions, including additional recognition of geographical objects and a wider range of attributes such as location and color (Fig. \ref{fig: align_examples}).

% \begin{figure}[H]
%   \centering
%   \includegraphics[width=\linewidth]{radar.png}
%   \caption{Overall performance comparison between \MODELNAME~and other MLLMs on different RS image understanding benchmarks.}
%   \label{fig: radar}
%   % \vspace{-0.5cm}
% \end{figure}

To enhance the model’s spatial awareness, we extend the LHRS-Instruct dataset~\citep{muhtar2024lhrs} with conversations that primarily focus on localization and perception.
Furthermore, we integrate an off-the-shelf robust visual instruction dataset that includes abundant negative samples, helping to balance the dataset and reduce the occurrence of hallucinations~\citep{liu2023LRV}.
Considering the necessity of a vision-centric~\citep{tong2024cambrian1fullyopenvisioncentric} design for holistic visual understanding, we scale up the vision encoder to accommodate inputs with larger resolutions. Additionally, we propose a novel bridge layer using the MoE architecture~\citep{jiang2024mixtralexperts} to further extend the model’s capacity, enabling lossless compression of visual information and dynamic mapping to the language domain.
With enhanced instructional data and an optimized modeling architecture, \MODELNAME~showcases exceptional spatial recognition capabilities while significantly minimizing the risk of hallucinations.

Finally, we conduct a thorough evaluation of various general-purpose and RS-specific MLLMs, not only on standard RS tasks such as classification, visual question answering, and visual grounding, but also on a multiple-choice question (MCQ) evaluation benchmark, LHRS-Bench~\citep{muhtar2024lhrs}, which is designed to comprehensively assess MLLMs in the RS domain. 
This facilitates a holistic assessment of instruction-following abilities and other RS-specific capabilities across multiple dimensions, such as perception and spatial awareness.

The main contributions of this study are:
\begin{enumerate}
    \item We propose a large-scale RS image-caption dataset, LHRS-Align-Recap, with high-quality captions generated through feature-guided image captioning. Additionally, we enlarge our instruction dataset by generating more conversations that emphasize spatial recognition and robustness.
    \item We scale up the vision encoder for higher-resolution inputs and design an MoE-based bridge layer to enhance model capacity. This enables more efficient compression of visual information with limited vision tokens, thereby improving language-vision alignment performance and visual understanding.
    \item We introduce a RS-specialized MLLM, \MODELNAME, and systematically evaluate its performance across a range of tasks to assess the fundamental task-solving capabilities. 
    % Additionally, we evaluate the effectiveness in complex RS perception scenarios and instruction-following using an MCQ dataset. This comprehensive evaluation provides a deeper understanding of the reliability of MLLMs as task solvers and offers valuable insights for future improvements.
    Additionally, our comprehensive evaluation through an MCQ dataset provides a deeper understanding of the reliability of MLLMs as task solvers and offers valuable insights for future improvements.
\end{enumerate}

\section{Related works}
\label{sec:rw}
\subsection{MLLM development in the RS community}
\label{subsec2.1}
MLLMs, such as GPT-4V~\citep{2023GPT4VisionSC} and Gemini~\citep{team2023gemini}, serve as versatile assistants with advanced capabilities in visual comprehension, reasoning, and human interaction. Their success has also fueled the development of RS-specific MLLMs designed for enhanced RS image interpretation~\citep{li2024vision, zhou2024towards}. RSGPT~\citep{hu2023rsgpt} is the first study to incorporate LLMs for RS visual-language tasks, followed by the creation of various MLLMs~\citep{kuckreja2024geochat, zhang2024earthgpt, zhan2024skyeyegpt} specifically trained with instruction tuning datasets generated from RS images. Beyond that,~\cite{muhtar2024lhrs} introduced LHRS-Bot, which leverages a newly proposed large-scale RS vision-language dataset and enhances visual features with a novel vision perceiver. \cite{pang2024h2rsvlm} proposed H2RSVLM, featuring enhanced self-awareness capabilities within the MLLM. \cite{luo2024skysensegpt} developed SkySenseGPT, supported by a large-scale instruction tuning dataset that features complex scenes. Collectively, these studies have significantly advanced the application of MLLMs in the RS community. However, current RS MLLMs still face several challenges, including a shortage of large-scale vision-language datasets with high-quality image captions, limitations in spatial recognition, a tendency toward hallucinations, and a lack of comprehensive performance evaluation—issues that this study aims to address and improve.

\subsection{Large-scale RS vision-language dataset}
\label{subsec2.2}
A diverse and extensive dataset is the most crucial element for effective vision-language model training. Additionally, the quality of captions plays a significant role in aligning the vision and language modalities in MLLMs~\citep{nguyen2024improving, chen2023sharegpt4v, li2024recap-datacomp}. Several large-scale RS image-caption datasets have been introduced to improve multimodal alignment in RS scenarios.
RS5M~\citep{zhang2023rs5m} is the first extensive RS image-caption dataset, but it is assembled from web-crawled data, which is often noisy and lacks informativeness~\citep{nguyen2024improving}. 
SkyScript~\citep{wang2024skyscript} comprises accurately paired images and captions, but the captions are generated using simple rules, resulting in a lack of semantic richness, which is essential for effectively training MLLMs~\citep{wang2024skyscript}.
To address these issues, LHRS-Align~\citep{muhtar2024lhrs} generates captions based on OSM features by employing an LLM. However, these captions lack sentence diversity and may focus on salient objects, which is sub-optimal for modality alignment~\citep{chen2023sharegpt4v}. In this study, we explore feature-guided image recaptioning~\citep{chen2023sharegpt4v, li2024recap-datacomp} to enhance the quality of vision-language pretraining datasets.

% \subsection{MLLM Evaluation}
% \label{subsec2.3}
% Given the vast corpora of current MLLMs, reference-based metrics like BLEU, METEOR, and CIDEr are often inadequate for evaluating their performance~\citep{hessel2021clipscore}. CV always take measures like... In RS field, ... 

\section{Dataset}
\label{sec:dataset}

\subsection{LHRS-Align-Recap: feature-guided image recaptioning for multimodal alignment}
\label{subsec3.1}

\begin{figure}[t]
  \centering
  \includegraphics[width=\linewidth]{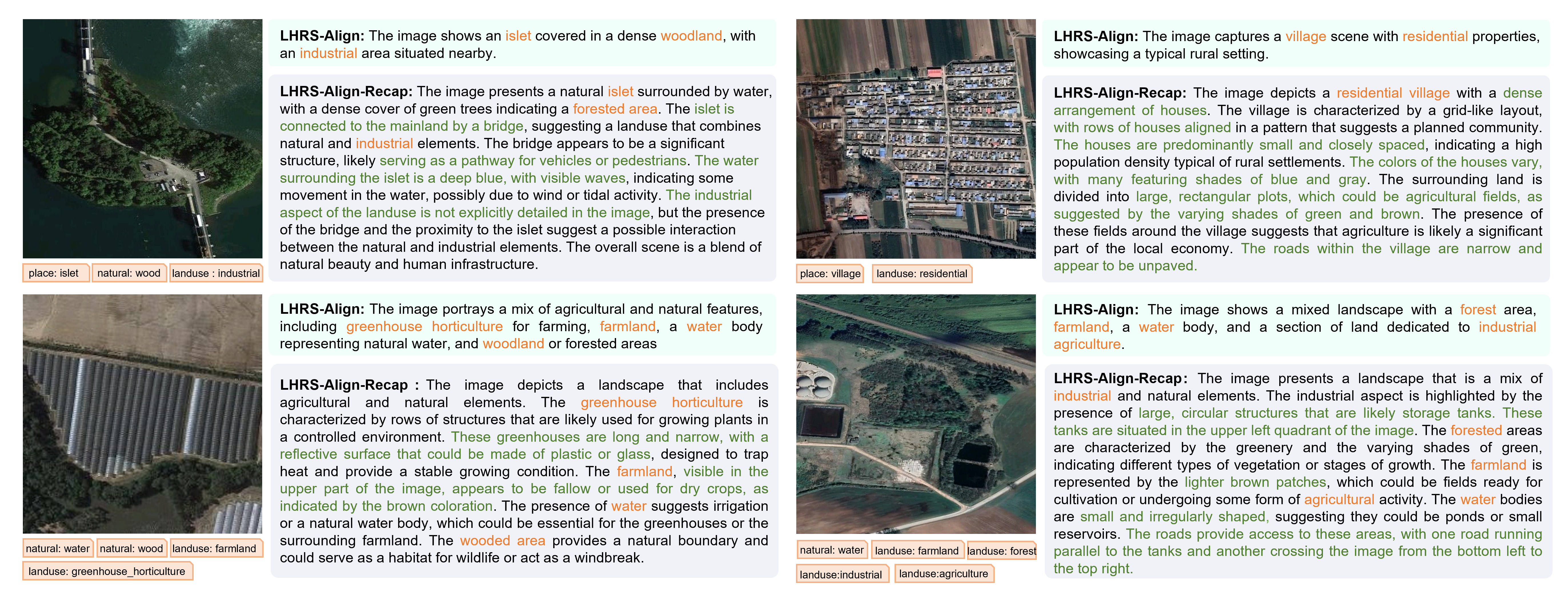}
  \caption{Four examples from LHRS-Align and LHRS-Align-Recap.}
  \label{fig: align_examples}
  % \vspace{-0.5cm}
\end{figure}

High-quality vision-language alignment datasets play a pivotal role for training a robust multimodal model~\citep{nguyen2022quality, fang2022data, gadre2024datacomp, xu2023demystifying}. To enhance vision-language alignment for RS-specific multimodal models, LHRS-Align~\citep{muhtar2024lhrs} was proposed by pairing RS images with OSM features, which are then used for generating captions with an LLM. 
Though effective for improving vision-language alignment, the captions in LHRS-Align are often concise and grammatically monotonous, and focus on key objects constrained by OSM features, as shown in Fig. \ref{fig: align_examples}, leading to suboptimal alignment across modalities~\citep{chen2023sharegpt4v}. 
To further enhance caption quality for more robust multimodal alignment, we leverage vision-language models for recaptioning and propose a new large-scale RS vision-language dataset, LHRS-Align-Recap.
Specifically, we prompt the Share-Captioner~\citep{chen2023sharegpt4v} to generate captions based on RS images and their corresponding OSM features from the LHRS-Align dataset~\citep{muhtar2024lhrs}, using the prompt design shown in Table \ref{fig: captioner prompt}.
In contrast to LHRS-Align~\citep{muhtar2024lhrs}, where captions are generated by a blind LLM, LHRS-Align-Recap is produced with an MLLM that can actually see the image, resulting in highly descriptive captions with more detail and diverse sentence structures, as illustrated by the examples in Fig. \ref{fig: align_examples}.

\begin{figure}[t]
  \centering
  \includegraphics[width=\linewidth]{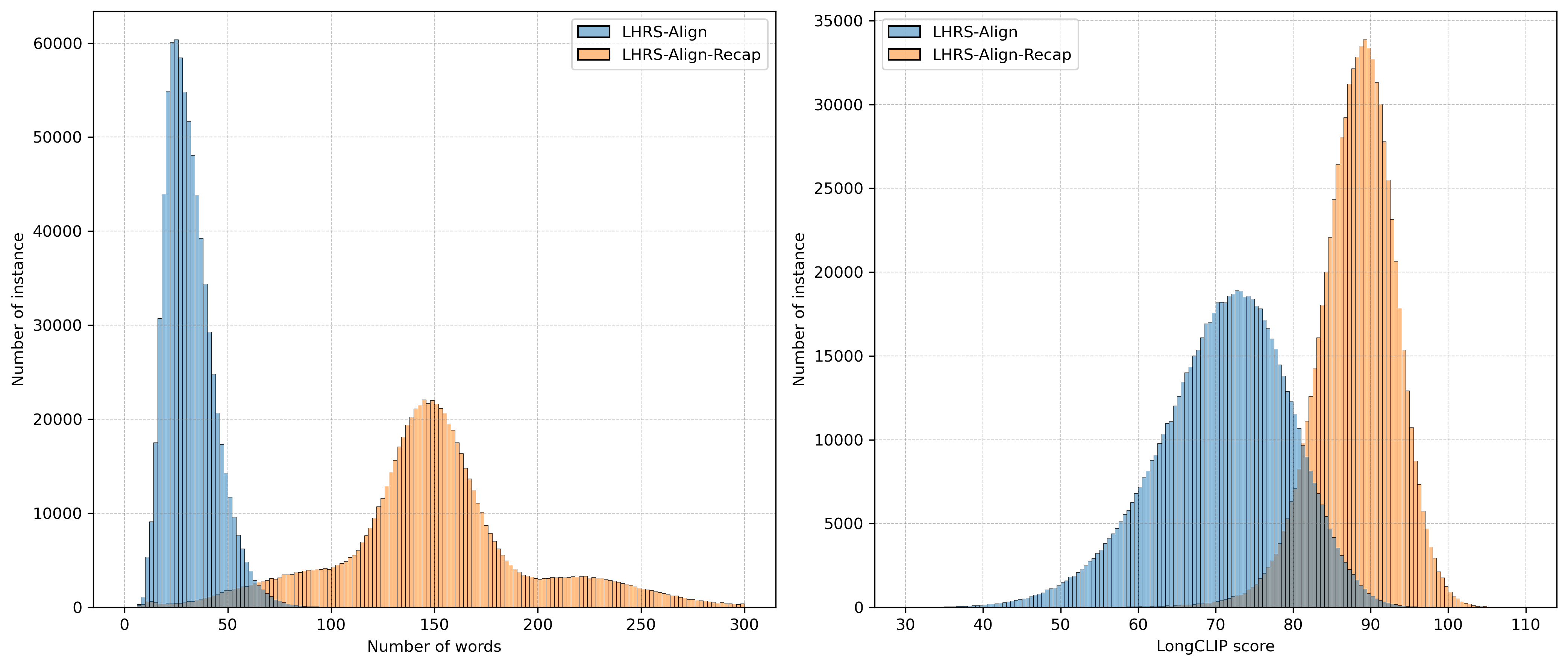}
  \caption{Caption length (left) and image-caption LongCLIP score (right) distributions for the LHRS-Align and the proposed LHRS-Align-Recap dataset.}
  \label{fig: align plot}
  % \vspace{-0.5cm}
\end{figure}

To gain a deeper understanding of the improvement, we conduct a statistical analysis of the original and new captions, focusing on two key aspects: the inherent distribution of the captions and the alignment quality of the captions. 
From the first perspective, we calculate the number of unique words and unique trigrams for both versions of the captions. These metrics gauge vocabulary richness and the structure and repetitive patterns, respectively.
As shown in Table \ref{tab:dataset stat}, the number of unique words and trigrams in LHRS-Align-Recap is nearly twice that of LHRS-Align, indicating increased vocabulary diversity and more varied sentence structures and phrase constructions. Additionally, the new captions have an average sentence length of 150 words, which is nearly five times that of the original captions, with a more even distribution, as shown in Fig. \ref{fig: align plot}. 
In terms of alignment quality, we use the CLIP score computed from LongCLIP~\citep{zhang2024longclip} to assess the semantic alignment between captions and images.
The higher CLIP score computed from the LHRS-Align-Recap dataset, as shown in Fig. \ref{fig: align plot} and Table \ref{tab:dataset stat}, demonstrates its greater effectiveness for vision-language alignment~\citep{nguyen2024improving}. A detailed quantitative validation of the improved dataset will be discussed in Section \ref{subsec:Ablation}.

\begin{table*}[!ht]
    \centering
    \begin{minipage}{0.99\columnwidth}
    \vspace{0mm}    
    \centering
    \caption{Prompt for generating captions with Share-Captioner based on RS images and OSM features.}
    \label{fig: captioner prompt}
    \begin{tcolorbox} 
        \centering
        \fontsize{8pt}{10pt}\selectfont
        \hspace{-6mm}
        \begin{tabular}{p{0.99\columnwidth}}
        Given the following image and its associated key-value features, generate a concise and descriptive caption that captures the essence of the image. The caption should reflect the relationships, context, and any notable details highlighted by the features. Ensure that the caption is coherent and informative, making use of the provided tags to enhance accuracy and detail.
        
        \VarSty{\{image\}}

        \VarSty{\{Key-Value Features\}}
        \end{tabular}
    \end{tcolorbox}
    \vspace{-2mm}
    \end{minipage}
\end{table*}

\begin{table}[t!]
\vspace{0.5cm}
\caption{Caption improvements in LHRS-Align-Recap compared to the original LHRS-Align, highlighting greater vocabulary richness, more varied sentence structures, and stronger alignment with images.}
\centering
\label{tab:dataset stat}
\scalebox{0.66}
{
    \begin{tabular}{lcccc}
    \toprule \rowcolor{mygray}
    \multicolumn{1}{c}{Dataset}    & No. of unique words     & No. of unique trigrams  & Average sentence length  & Average LongCLIP score    \\ \midrule

    LHRS-Align   & 8436      & $1.12×10^6$     & 31    & 70.81          \\
    LHRS-Align-Recap   & 15345     & $2.64×10^6$     & 150   & 88.12          \\ \bottomrule
    \end{tabular}
}
\end{table}

\subsection{Instruction tuning datasets for training \MODELNAME}
\label{subsec3.2}
Visual instruction fine-tuning~\citep{liu2024visual} plays a significant role in MLLM training by enabling LLMs to better understand visual features and follow human instruction~\citep{xu2024vision-flan}. 
This process enhances the model's capabilities and controllability, allowing it to generalize to diverse tasks and provide feedbacks that align with human preferences~\citep{li2023vision, zhang2023instruction}. 
We introduce four datasets used for instruction tuning LHRS-Bot-Nova as below, including a multi-task dataset, LHRS-Instruct, LRV-Instruct, and the proposed LHRS-Instruct-Plus~(Table \ref{tab:sft dataset}).

\begin{table}[H]
\caption{Instruction tuning datasets for training \MODELNAME.}
\centering
\label{tab:sft dataset}
\resizebox{\columnwidth}{!}
{
    % \begin{tabular}{|l|l|l|l|}
    \begin{tabular}{lllll}
    \toprule 
    \rowcolor{mygray}
    Dataset Name                & Original dataset type      & Original dataset          & Instance num. & Instruction task        \\ \midrule
    \multirow{8}{*}{Multitask}   & \multirow{2}{*}{VQA}       & RSVQA-HR~\citep{lobry2020rsvqa}      & 213160 & \multirow{2}{*}{VQA with consize answers}     \\ 
                                 &                            & RSVQA-LR~\citep{lobry2020rsvqa}      & 11440 &                          \\ \noalign{\vskip 2pt} \cline{2-5} \noalign{\vskip 2pt}
                                 & \multirow{3}{*}{Classification}       & UCM~\citep{ucm}         & 2100  & \multirow{3}{*}{Classification} \\ 
                                 &                            & fMoW~\citep{christie2018fmow}      & 5352    &                          \\
                                 &                            & METER-ML~\citep{zhu2022meter}     & 1400 &                          \\ \noalign{\vskip 2pt} \cline{2-5} \noalign{\vskip 2pt}
                                 & \multirow{2}{*}{Visual grounding}    & RSVG~\citep{sun2022rsvg}   & 2428        & \multirow{2}{*}{Visual grounding} \\ 
                                 &                            & DIOR-RSVG~\citep{zhan2023dior-rsvg}  & 14030   &                          \\ \noalign{\vskip 2pt} \cline{2-5} \noalign{\vskip 2pt}
                                 &  Image captioning         & RSICD~\citep{lu2017rsicd}    & 1000     &  Briefly image captioning                  \\ \midrule
    \multirow{3}{*}{LHRS-Instruct} & \multirow{3}{*}{Image captioning} & NWPU~\citep{cheng2022nwpucap}         & 111755           & Conversation             \\ 
                                 &                            & RSITMD~\citep{rsitmd}              & 12200       & Conversation             \\
                                 &                            & LHRS-Align~\citep{muhtar2024lhrs}    & 29671    & Conversation, Detailed description, Visual reasoning\\ \midrule
    \multirow{2}{*}{LHRS-Instruct-Plus} & \multirow{2}{*}{Object detection} & DOTAv2~\citep{ding2021dotav2}   & 67143      & Conversation, Object detection \\ 
                                 &                            & FAIR1M~\citep{sun2022fair1m}                & 253446     & Conversation, Object detection \\ \midrule
    LRV-Instruct                 & Instruction tuning         & LRV-Instruct~\citep{liu2023LRV}             & 140974             & Robust Visual Instruction with negative samples \\ \bottomrule
    \end{tabular}
}
\end{table}

We utilize the multi-task instruction dataset and the LHRS-Instruct dataset proposed in \citep{muhtar2024lhrs} to enhance task-solving and complex understanding capabilities.
The former was constructed by combining various public RS datasets with manually created instruction templates, and the latter was created by prompting LLMs to create complex conversations using selected samples from RS caption datasets.

To further improve the ability of \MODELNAME~for understanding spatial relationships, we construct a novel instruction dataset, LHRS-Instruct-Plus. 
Specifically, the proposed dataset is generated by prompting GPT-4V to generate various conversations from the two RS object detection datasets, DOTAv2~\citep{ding2021dotav2} and FAIR1M~\citep{sun2022fair1m}. 
The prompt we used is shown in Table \ref{tab: instruct prompt}. The generated conversations primarily includes object identification, as well as other tasks such as object counting, image description, and visual reasoning. 

Furthermore, recent studies have shown that current MLLMs tend to answer ``Yes" to any given instruction, even when the correct response should be ``No", a phenomenon known as hallucination~\citep{li2023eval-hall, bai2024hallucination, liu2023LRV}. One reason for this issue is the imbalance in instruction datasets, where positive instructions predominate and the importance of negative samples is overlooked. 
To mitigate this problem, we integrate the LRV-Instruct dataset~\citep{liu2023LRV}, which contains both positive and negative instructions, into our instruction datasets, for more robust visual instruction tuning. This makes the trained model more reliable and enhances its ability to accurately differentiate between affirmative and negative responses, thereby reducing the occurrence of hallucinations.

\begin{figure}[H]
  \centering
  \includegraphics[width=\linewidth]{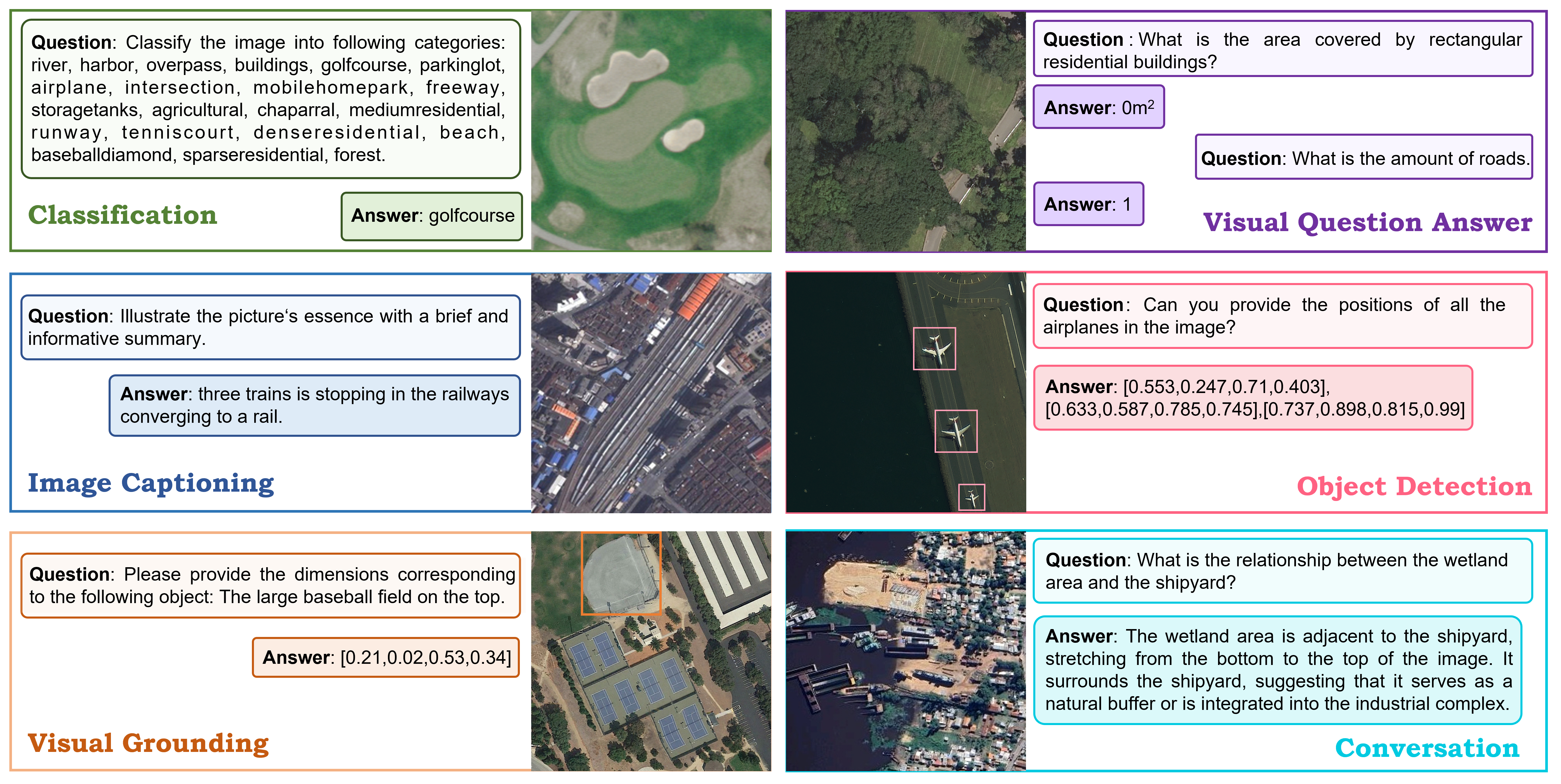}
  \caption{Six examples of various instruction tasks from the instruction datasets used to train \MODELNAME.}
  \label{fig: instruction examples}
  % \vspace{-0.5cm}
\end{figure}

In general, all the instruction tuning datasets for training our \MODELNAME~are listed in Table \ref{tab:sft dataset}, and some examples are shown in Fig. \ref{fig: instruction examples}.

\begin{table*}[!ht]
    \centering
    \begin{minipage}{0.99\columnwidth}
    \vspace{0mm}    
    \centering
    \caption{Prompt for generating LHRS-Instruct-Plus with GPT-4V based on RS object detection datasets.}
    \label{tab: instruct prompt}
    \begin{tcolorbox} 
        \centering
        \fontsize{8pt}{10pt}\selectfont
        \hspace{-6mm}
        \begin{tabular}{p{0.99\columnwidth}}
                \VarSty{System Message:}
                
                    You are an AI visual assistant that can design a conversation for analysing images. You will be presented with an image along with the category of the main object and the coordinates of its bounding box, given as (x1, y1, x2, y2), with values ranging from 0 to 1. These coordinates correspond to the top-left x, top-left y, bottom-right x, and bottom-right y positions, respectively. Your task is to design conversations about the image based on the object's categories and their locations. The dialogues should primarily focus on the number of objects, their locations, and the spatial relationships between them. 
                    When designing conversation, use the information from the coordinates naturally, without mentioning that the source is the given bounding box.
                    
                \vspace{2mm}
                \VarSty{In-Context Example:}
                
                \VarSty{User:}
                \hspace{12cm}\multirow{5}{*}{ \includegraphics[height=2.0cm]{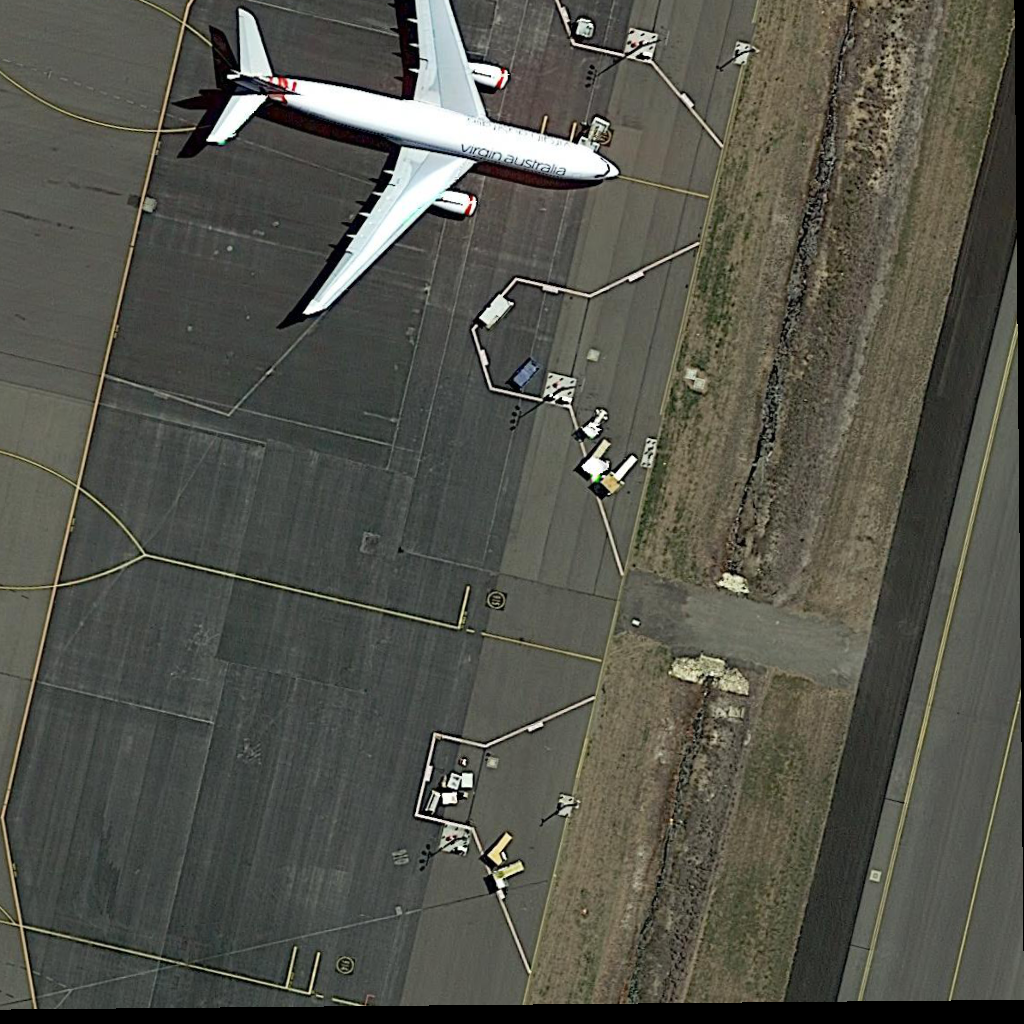}}  \\
                
                airplane [0.399,0.163,0.452,0.293]
                
                \VarSty{Assistant:}
                
                \textbf{Questions:}
                
                How many airplane can you see?
                
                \textbf{Answer:}

                I can see an airplane in the image.

                \textbf{Questions:}
                
                Where is the airplane positioned in the image?
                
                \textbf{Answer:}

                The airplane is mostly in the middle of the image but slightly tilted towards the right. Its tail end is closer to the center, while the nose is pointing slightly upwards and to the right.

                \textbf{Questions:}
                
                Where is the airplane positioned in the image?
                
                \textbf{Answer:}

                The airplane is mostly in the middle of the image but slightly tilted towards the right. Its tail end is closer to the center, while the nose is pointing slightly upwards and to the right.

                \textbf{Questions:}
                
                [DET] Please given me the exact position of the airplane.
                
                \textbf{Answer:}

                $<$bbox$>$[0.399,0.163,0.452,0.293]$<$/bbox$>$
        \end{tabular}
    \end{tcolorbox}
    \vspace{-2mm}
    \end{minipage}
\end{table*}

\section{Methodology}
\label{sec:method}
\MODELNAME~incorporates an enhanced vision encoder and a new vision perceiver with an MoE structure for improved vision-language alignment. 
In this section, we delve into each component of \MODELNAME, detailing how the enhanced architecture achieves better vision-language alignment. Subsequently, we explain the curriculum training strategy used for training \MODELNAME.

\subsection{Model architecture}
\begin{figure}[t!]
  \centering
  \includegraphics[width=\linewidth]{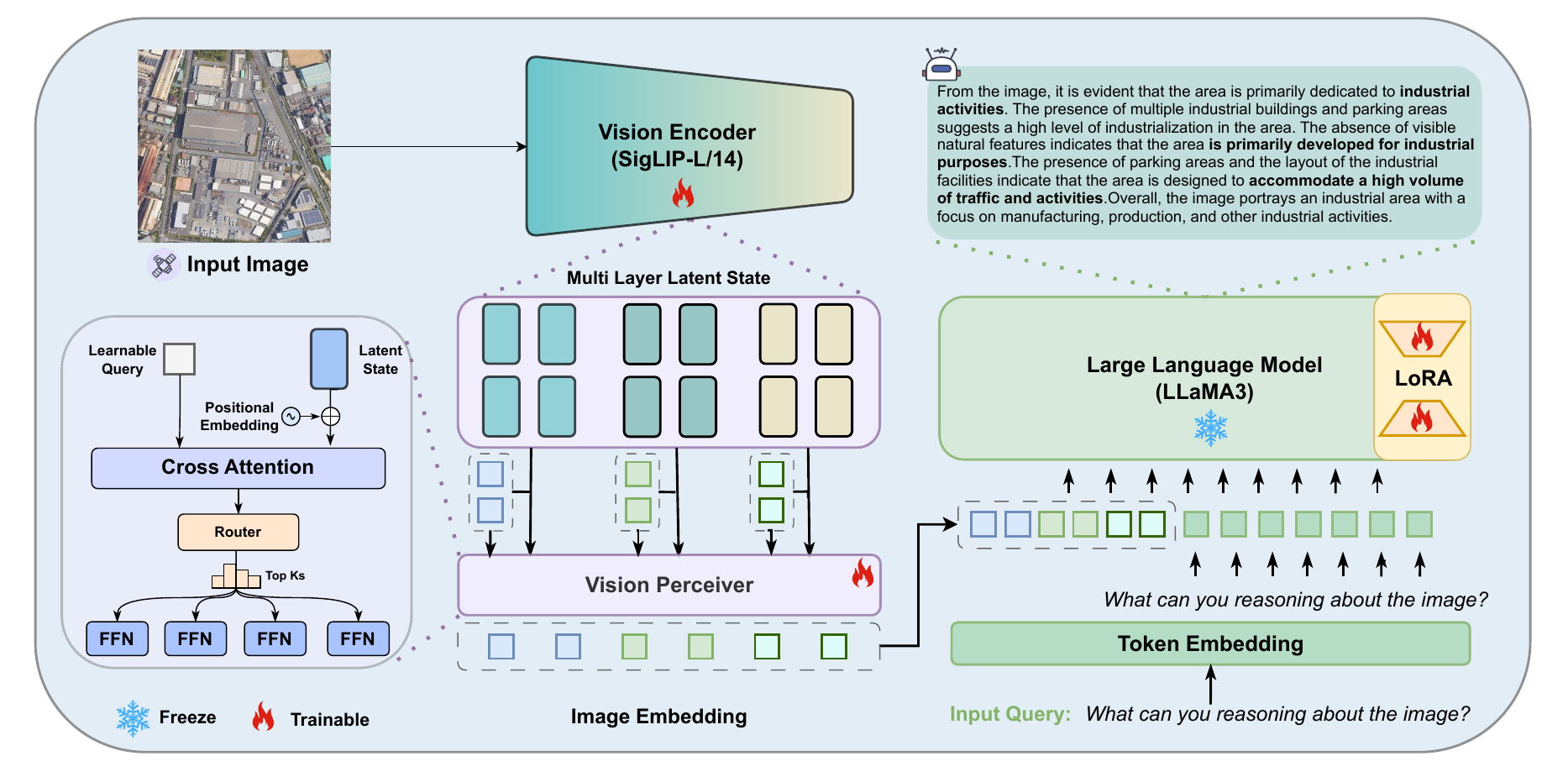}
    \caption{Architecture of LHRS-Bot-Nova. LHRS-Bot-Nova employs learnable queries in conjunction with an MoE vision perceiver to summarize multi-level visual representations, which are then concatenated with language token embeddings as input to the LLM.}
  \label{fig:architecture}
\end{figure}

\MODELNAME~is primarily composed of three components: a vision encoder, a vision perceiver, and a foundational LLM. The overall architecture of \MODELNAME~is presented in Fig. \ref{fig:architecture}. 

\textbf{Vision Encoder.} Recent studies have highlighted that increasing the input image resolution enhances the visual understanding ability of MLLMs~\citep{tong2024cambrian1fullyopenvisioncentric,liu2024llava1.5,li2024llavaonevisioneasyvisualtask}. 
Therefore, we adopt SigLIP-L/14~\citep{zhai2023sigmoidlosslanguageimage} with an input resolution of 336×336 as the vision encoder to extract more detailed visual signals.
Additionally, we follow the strategy in~\cite{muhtar2024lhrs} to extract multi-level visual information, providing additional visual supervision for more data-efficient vision-language alignment.

\textbf{Vision Perceiver.}
Considering the additional computational and memory overhead associated with the extra vision tokens from multi-layer visual signals, we follow~\cite{muhtar2024lhrs} to summarize different layers of visual signals using learnable query-based cross-attention. 
Additionally, we adopt a decreasing query allocation strategy to manage the higher redundancy in deeper levels of the vision hidden states within the vision encoder~\citep{bolya2022token}.
However, while this design effectively compresses vision tokens by reducing their number, it may lead to a loss of visual details~\citep{tong2024cambrian1fullyopenvisioncentric}. 
Therefore, inspired by the observation that the feed-forward network (FFN) layer acts as network memory~\citep{geva2021transformerfeedforwardlayerskeyvalue}, we incorporate MoE architecture~\citep{jiang2024mixtralexperts} into each layer of the FFN in the Vision Perceiver to expand the model's memory capacity, allowing for the extraction of more detailed visual information even with fewer vision tokens.
Specifically, given the learnable query $\mathbf{Q}^i \in \mathbb{R}^{n^i \times d}$ and the vision token $\mathbf{X}^i \in \mathbb{R}^{L \times d}$, where the superscript $i$ denotes the $i$-th level (with $i \in {1, 2, 3}$ typically corresponding to low, medium, and deep levels), $L$ denotes the number of tokens, and $d$ is the hidden dimension, we first summarize $\mathbf{X}^i$ using $\mathbf{Q}^i$ through cross-attention:
\begin{equation}\label{eq:cross_attn}
    \mathbf{h}^i = \text{Softmax}
    \left(
        \frac{\mathbf{Q}^i(\mathbf{W}_k\mathbf{X}^i + p)^T}{\sqrt{d}}    \right)
        (\mathbf{W}_v\mathbf{X}^i + p) 
        \in \mathbb{R}^{n^i\times d},
\end{equation}
where $\mathbf{W}_k, \mathbf{W}_v \in \mathbb{R}^{d \times d}$ are learnable parameters, $p$ denotes sinusoidal positional embedding, and $\mathbf{h}_i$ represents the summarized vision tokens for the $i$-th level.
Then, we concatenate all levels of summarized results, $\mathbf{h} := [\mathbf{h}^1;\mathbf{h}^2;\mathbf{h}^3] \in \mathbb{R}^{(n^1+n^2+n^3) \times d}$. For each token $\mathbf{h}_t$ in the concatenated result, we compute the output $\mathbf{O}_t$ of the MoE-FFN layer as follows:
\begin{equation}\label{eq:moe_eq1}
    \mathbf{O}_t = \mathbf{h}_t + \sum_{j=1}^{N_e}\textsl{g}_{j,t}\text{FFN}_{j}(\mathbf{h}_t)
\end{equation}
\begin{equation}\label{eq:moe_eq2}
    \textsl{g}_{j,t} = 
    \begin{cases}
        s_{j,t}, & s_{j,t} \in \text{Top}K(\{s_{k,t}|1\leq k \leq N_e\}, K), \\
        0,       & \text{otherwise},
    \end{cases}
\end{equation}
\begin{equation}
    s_{j,t}=\text{Softmax}_j(\mathbf{h}_t^T\mathbf{W}_{\text{router}}),
\end{equation}
where $N_e$ is the number of experts, $\text{FFN}_j(\cdot)$ denotes the $j$-th expert, $K$ represents the number of activated routed experts for each token, $\textsl{g}_{j,t}$ is the gate value for the $j$-th expert, $s_{j,t}$ is the token-to-expert affinity, $\mathbf{W}_{\text{router}}$ is the learnable parameter for the expert router, and Top$K(\cdot,K)$ denotes the set comprising the $K$ highest scores among the affinity scores calculated for the $t$-th token across all routed experts.
With this MoE architecture in the FFN layer and the per-token sparse dynamic routing strategy, we expand the model's memory capacity without introducing additional inference activation, allowing for the extraction and retention of more detailed visual information.

\textbf{Large Language Model.} 
We employ the improved LLaMA3-8B~\citep{dubey2024llama3} architecture as the central ``brain" of \MODELNAME, enabling it to interpret various signals from vision and language to respond to given instructions.

\subsection{Training strategy}
We follow the three-stage curriculum learning strategy introduced by \cite{muhtar2024lhrs}, which includes pre-training, multi-task instruction fine-tuning, and supervised fine-tuning stages.

In the pre-training stage, we use the LHRS-Bot-Recap dataset to train the vision encoder and vision perceiver, mapping multi-level visual signals into the language domain.

During the multi-task instruction fine-tuning stage, we unfreeze the LLM with low-rank adapters (LoRA)~\citep{hu2021loralowrankadaptationlarge} and fine-tune both the vision perceiver and LoRA using the multi-task instruction dataset to improve the multimodal task-solving capabilities of \MODELNAME.

Finally, in the supervised fine-tuning stage, we use all the instruction data from the LHRS-Instruct, LHRS-Instruct-Plus, and LRV-Instruct~\citep{liu2023LRV} datasets to further train the vision perceiver and LoRA adapter, enhancing the conversational and reasoning capabilities of \MODELNAME.

\section{Experiments}
\label{sec:experiments}

\subsection{Experimental setup}
\label{subsec:setup}

\textbf{Evaluation benchmarks.} We evaluate our model in various RS image understanding benchmarks.

\textbf{1) Scene classification.} We use the test sets of seven datasets: AID~\citep{xia2017aid}, WHU-RS19~\citep{Dai2011WHURS19}, SIRI-WHU~\citep{SIRI-WHU}, EuroSAT~\citep{helber2019eurosat}, NWPU~\citep{NWPU}, METER-ML~\citep{zhu2022meter}, and fMoW~\citep{christie2018fmow}. The first four datasets are evaluated in a zero-shot setting.

\textbf{2) Visual question answering (VQA).} We utilize two datasets: the test sets of RSVQA-HR and RSVQA-LR~\citep{lobry2020rsvqa}.

\textbf{3) Visual grounding.} We use two datasets: the test sets of RSVG~\citep{sun2022rsvg} and DIOR-RSVG~\citep{zhan2023dior-rsvg}.

\textbf{4) Benchmarks for RS MLLMs.} LHRS-Bench~\citep{muhtar2024lhrs} with MCQs is used for systematic evaluation of MLLMs in RS image understanding.

\textbf{Baselines.} We evaluate \MODELNAME~against the 7B variants of several powerful open-source general domain MLLMs, including LLaVA-1.5~\citep{liu2024llava1.5}, MiniGPTv2~\citep{chen2023minigptv2}, InstructBLIP~\citep{instructblip}, mPLUG-Owl2~\citep{ye2024mplug2}, QWen-VL-Chat~\citep{Qwen-VL}, and InternLM-Xcomposer~\citep{zhang2023internlm}, across various tasks. 
For comparisons with RS MLLMs, we use the accuracies reported in the papers whenever possible, including GeoChat~\citep{kuckreja2024geochat}, SkyEyeGPT~\citep{zhan2024skyeyegpt}, H2RSVLM~\citep{pang2024h2rsvlm}, and SkySenseGPT~\citep{luo2024skysensegpt}.

\textbf{Implementation details.}
We extract the latent states at layers
$\{N_{L}/3, 2N_{L}/3, N_{L}-1\}$ for summarization through the MoE vision and LHRS-Bot~\citep{muhtar2024lhrs} perceiver, where $N_L$ denotes the number of layers in the vision encoder. The allocated queries for each layer are \{112, 96, 64\}.
The vision perceiver implementation comprises six layers of cross-attention and FFN, with each layer containing four FFN experts, and the number of activated experts $K$ in Equation~\ref{eq:moe_eq2} is set to 2.
We apply the LoRA module to every linear layer of the LLM, setting the rank $r$ and the scale factor $\alpha$ for LoRA to 128 and 256, respectively. Additionally, we introduce task identifiers [CLS], [CONCISE], and [DET] for classification, concise vision-language answering, and visual 
grounding tasks, respectively. 
All 3 stages of training were done with AdamW optimizer on 8 $\times$ H100 GPUs for 1 epoch, the hyperparameters used at each stage are presented in Table~\ref{tab:hyper}.

% \begin{wrapfigure}{r}{0.5\textwidth}
%     \begin{minipage}{0.5\textwidth}
%         \centering  
%         \setlength{\extrarowheight}{-4pt}
%         \captionof{table}{Hyperparameter settings for each stage.}
%         \label{tab:hyper}
%         \scalebox{0.78}
%         {
%             \begin{tabular}{@{}lccc@{}}
%             \toprule \rowcolor{mygray}
%                                  & Stage1  & Stage2     & Stage3      \\ \midrule 
%             Learning Rate        & 0.0002  & 0.0001     & 0.0001      \\
%             Global Batch Size    & 128     & 64         & 64           \\
%             Weight Decay         & 0.      & 0.01       & 0.01            \\
%             $\beta_1$             &  \multicolumn{3}{c}{0.9}  \\
%             $\beta_2$             &  \multicolumn{3}{c}{0.95} \\
%             Gradient Norm        & \multicolumn{3}{c}{1.0}               \\
%             Scheduler            & \multicolumn{3}{c}{Cosine}         \\
%             Warmup Steps         & 300     & 100        & 0          \\ \bottomrule
%             \end{tabular}
%         }
%         \vspace{-5mm}
%     \end{minipage}
% \end{wrapfigure}

\begin{figure}[H]
    \centering  
    \setlength{\extrarowheight}{-4pt}
    \captionof{table}{Hyperparameter settings for each stage.}
    \label{tab:hyper}
    \scalebox{0.85}
    {
        \begin{tabular}{@{}lccc@{}}
        \toprule \rowcolor{mygray}
                             & Stage1  & Stage2     & Stage3      \\ \midrule 
        Learning Rate        & 0.0002  & 0.0001     & 0.0001      \\
        Global Batch Size    & 128     & 64         & 64           \\
        Weight Decay         & 0.      & 0.01       & 0.01            \\
        $\beta_1$             &  \multicolumn{3}{c}{0.9}  \\
        $\beta_2$             &  \multicolumn{3}{c}{0.95} \\
        Gradient Norm        & \multicolumn{3}{c}{1.0}               \\
        Scheduler            & \multicolumn{3}{c}{Cosine}         \\
        Warmup Steps         & 300     & 100        & 0          \\ \bottomrule
        \end{tabular}
    }
    \vspace{-5mm}
\end{figure}

\subsection{Quantitative results on RS image understanding}
\label{subsec:Quantitative}

\textbf{Scene classification.}
We evaluate the scene classification accuracy of \MODELNAME~compared with other open-source MLLMs, highlighting its broad knowledge in recognizing geographical features. As shown in Table \ref{tab:cls_compare}, the classification accuracy of \MODELNAME~surpasses other general MLLMs by a significant margin, thanks to the RS domain-specific training. Notably, compared to LHRS-Bot, the classification performance has improved across nearly all datasets, with a significant overall accuracy increase of up to 4.77\%, demonstrating the powerful scene understanding ability of \MODELNAME. 
Since the AID, WHU-RS19, SIRI-WHU, and EuroSAT datasets were entirely absent from the multi-task training data, the accuracies achieved on these datasets reflect a zero-shot setting, demonstrating the strong generalization capability of \MODELNAME. 
% Overall, the high classification accuracy underscores the effectiveness of optimizing the large-scale vision-language alignment dataset.

\begin{table}[H]
\caption{Accuracies (\%) of different MLLMs on various scene classification datasets.}
\vspace{-0.2cm}
\label{tab:cls_compare}
\resizebox{\columnwidth}{!}{%
\begin{tabular}{lcccccccc}
\toprule 
\rowcolor{mygray}
\multicolumn{1}{l}{Method} & AID & WHU-RS19 & SIRI-WHU & EuroSAT & NWPU & METER-ML & fMoW     & Avg.  \\ \midrule
LLaVA-1.5                & 31.10 & 54.55    & 17.71    & 26.12   & 34.96   & 21.73    & 11.43  & 28.23 \\
MiniGPTv2                & 32.96 & 64.80    & 35.46    & 38.56   & 28.15   & 14.29    & 5.20   & 31.35 \\ 
InstructBLIP             & 29.50 & 36.76    & 18.20    & 20.25   & 34.01   & 14.42    & 6.71   & 22.84 \\
mPLUG-OWL2               & 48.79 & 72.66    & 54.83    & 33.29   & 46.58   & 36.27    & 17.85  & {44.32} \\
QWen-VL-Chat             & 55.30 & 72.25    & 54.58    & 26.42   & 42.73   & 38.77    & 6.89   & 42.42 \\
InternLM-XComposer       & 51.61 & 72.89    & 46.83    & 39.70   & 47.51   & 40.21    & 11.28  & 44,29 \\ 
LHRS-Bot              & \textbf{91.26}  & \underline{93.17}    & \underline{62.66}    & \underline{51.40}    & \underline{83.94}   & \underline{69.81}   & \underline{56.56} & \underline{71.83} \\
\rowcolor{cyan!50}
\MODELNAME      & \underline{88.32} & \textbf{95.63}    & \textbf{74.75}    & \textbf{63.54}    & \textbf{86.80}   & \textbf{70.05}   & \textbf{57.11} & \textbf{76.60} 
\\ \bottomrule
\end{tabular}
}
\vspace{-0.25cm}
\end{table}

\textbf{Visual question answering.}
We report the VQA results of \MODELNAME~on two RSVQA datasets by comparing with other general domain MLLMs and RS domain MLLMs in Table \ref{tab:vqa_compare}. It can be seen that the VQA results of \MODELNAME~surpass those of other general domain MLLMs by a significant margin. When compared to other RS-specific MLLMs, our model achieves comparable results on the RSVQA-LR data and demonstrates a significant advantage on the RSVQA-HR data. Overall, \MODELNAME~achieves the highest VQA accuracy, which slightly outperforms LHRS-Bot.

\begin{table}[H]
\vspace{0.5cm}
\caption{Accuracies (\%) of various general-domain and RS-specific MLLMs on visual question answering task in the RSVQA dataset.}
\vspace{-0.2cm}
\label{tab:vqa_compare}
\centering
\resizebox{\columnwidth}{!}{%
\begin{tabular}{l|cccc|ccc|c}
\hline \rowcolor{mygray}
Method & \multicolumn{4}{c|}{RSVQA-LR} & \multicolumn{3}{c|}{RSVQA-HR}  & Overall                                                             \\
                         & Rural/Urban & Presence & Compare & Avg.  & Presence  & Compare &  Avg.   &  Avg. \\ \hline
LLaVA-1.5                & 59.22       & 73.16    & 65.19   & 65.86 & 48.96    & 59.02   &  53.99   & 61.11 \\
MiniGPTv2                & 60.02       & 51.64    & 67.64   & 59.77 & 68.34    & 64.71   &  66.53   & 62.47 \\
InstructBLIP             & 62.62       & 48.83    & 65.92   & 59.12 & 62.63    & 62.90   &  62.77   & 60.58 \\
mPLUG-Owl2               & 57.99       & 74.04    & 63.69   & 65.24 & 47.60    & 58.47   &  53.04   & 60.36 \\
QWen-VL-Chat             & 62.00       & 47.65    & 66.54   & 58.73 & 61.75    & 65.98   &  63.87   & 60.78 \\ 
InternLM-XCompose        & 59.00       & 66.74    & 52.91   & 59.55 & 67.79     & 66.62  &  67.21   & 62.61 \\ \hline
% RSGPT                    & \textbf{94.00}       & \textbf{91.17}    & \underline{91.70}  & \textbf{92.29} & \underline{90.92}          & \underline{90.02}        &  \underline{90.47}    \\
GeoChat                  & \underline{91.09}  & \underline{90.33}  & \textbf{94.00} & \underline{91.81} & 58.45  & 83.19  & 70.82 & 83.41 \\
SkyEyeGPT                & 88.93       & 88.63    & 75.00  & 84.16  & 80.00    & 80.13   &  82.56   & 82.54 \\
H2RSVLM                  & 88.00       & 89.58    & 89.79  & 89.12  & 65.00  & 83.70    & 74.35     & 83.21  \\
SkySenseGPT              & \textbf{95.00} & \textbf{91.07}    & \underline{92.00}  & \textbf{92.69}  & 69.14   & 84.14  & 76.64 & 86.27 \\
LHRS-Bot                 & 89.07       & 88.51    & 90.00  & 89.19  & \textbf{92.57}  & \textbf{92.53}  & \textbf{92.55} & \underline{90.54}\\ \rowcolor{cyan!50} 
\MODELNAME          & 89.11       & 89.00     & 90.71  & 89.61 & \underline{91.68}   & \underline{92.44}   & \underline{92.06} & \textbf{90.59}   \\ \bottomrule
\end{tabular}
}
\vspace{-0.3cm}
\end{table}

\textbf{Visual grounding.}
The visual grounding accuracies for the RSVG and DIOR-RSVG datasets, using accuracy@0.5 as the evaluation metric, are presented in Table \ref{tab:vg_compare}. In comparison with other MLLMs, \MODELNAME~achieves the highest accuracies on both datasets. Compared to LHRS-Bot, our improved model demonstrates significantly better object interpretation and localization, as evidenced by a 6.58\% increase in average accuracy, which demonstrates the effectiveness of our data improvement.

\begin{table}[H]
\vspace{0.5cm}
\caption{Comparison of different MLLMs on visual grounding with the evaluation metric of accuracy@0.5 (\%), where a prediction is correct if the IoU between the predicted and ground-truth bounding boxes exceeds 0.5.}

\vspace{-0.2cm}
\centering
\label{tab:vg_compare}
\scalebox{0.8}
{
    \begin{tabular}{lccc}
    \toprule \rowcolor{mygray}
    \multicolumn{1}{l}{Method}  & RSVG      & DIOR-RSVG                  & Avg.  \\ \midrule
    QWen-VL-Chat                & 44.76     & 80.65                      & 62.71  \\
    MiniGPTv2                   & 46.64     & 85.99                      & 66.32  \\ \midrule
    SkyEyeGPT                   & {70.50}   & \underline{88.59}    & {79.55}   \\ 
    LHRS-Bot                    & \underline{73.45} & {88.10}    & \underline{80.78}  \\  
    \rowcolor{cyan!50} 
    \MODELNAME                  & \textbf{81.85} & \textbf{92.87}    &  \textbf{87.36}  \\   \bottomrule
    \end{tabular}
}
\end{table}

% \textbf{\textcolor{red}{Other evaluation.}}
% \textcolor{red}{TODO: Plan to add another dataset for evaluation.}

\subsection{Evaluation on LHRS-Bench}
\label{subsec:bench eval}

\textbf{Evaluation method.}
The evaluation of LLMs and MLLMs can be broadly categorized into generation-based and multiple-choice-based approaches. Generation-based methods involve scoring open-ended answers, typically requiring human or LLM for assessments~\citep{zheng2024judging}, which can introduce subjectivity. Consequently, many benchmarks employ MCQs to assess LLM and MLLM capabilities~\citep{MMLU, zhang2024multiple,liu2024mibench,li2024seed-BENCH}, which can quantitatively evaluate the model with an objective accuracy metric. In the RS field, MCQ benchmarks have also been introduced for RS MLLM evaluation~\citep{muhtar2024lhrs,luo2024skysensegpt}. In this study, we thoroughly explore how to effectively evaluate RS MLLMs using MCQ benchmarks, with the LHRS-Bench~\citep{muhtar2024lhrs} dataset as example.

% \textbf{Choice answering.}
% To obtain the model's answer, we initially applied a method that calculates the softmax over the output logits corresponding to the option token~\citep{zhang2024multiple,li2024seed-BENCH} , known as MULTIPLE CHOICE PROMPTING ~\citep{robinson2023leveraginglargelanguagemodels}. \textcolor{orange}{However, we found this method...}. Consequently, we opted to directly ask the model to generate the correct answer based on the given question and choices.

\textbf{Choice assignment.}
When directly prompting the MLLM to answer the correct choice, the model may sometimes ignore the instruction and output the full context of the candidate choices~\citep{liu2023mmbench}. To address this, \cite{liu2023mmbench} applies an exact matching approach between the model's output and the correct choice. However, this method often requires additional validation of ChatGPT-based matching because exact matching frequently fails. \cite{muhtar2024lhrs} employs substring matching, considering an answer correct if the correct context is a substring of the model’s output. However, this strategy can lead to mismatches: for example, if the correct context is ``industrial", an answer like ``residential but not industrial" might be mistakenly judged as correct. To resolve this, we use a simple approach: the prompt asks the model to ``Only answer with the letter corresponding to the given choices, such as A., B., etc.", and the answer is considered correct only if it strictly matches the expected letter\footnote{Interestingly, in practice, we find that all the models used for comparison can follow this simple prompt except gpt-4o-mini. Therefore, we use one-shot in-context learning to prompt it for a fair ability comparison.}. This metric can also evaluate the model's ability to follow instructions since a robust model should understand that the user is only interested in the correct letter.

\textbf{Robust evaluation.}
Recent studies reveal a phenomenon of selection bias and random guessing when evaluation with MCQs~\citep{myrzakhan2024Open-LLM-Leaderboard, wang2024Beyondtheanswers,robinson2023leveraginglargelanguagemodels}, where models tend to favor certain options or provide random answers when unable to solve a question. This leads to unreliable evaluation accuracies. To address this, techniques such as option reordering, increasing or reducing options, and altering them have been proposed~\citep{wang2024Beyondtheanswers}. In our experiments, we employ CircularEval~\citep{liu2023mmbench,liu2024mibench}, which feeds the same question to an MLLM multiple times while rotating the correct answer among the options and checks if the model consistently selects the correct answer across all attempts. We found that using this strategy is crucial for enhancing the robustness of the MCQ evaluation for RS MLLMs: without it, the accuracy on LHRS-Bench of various MLLMs could increase by around 20\%, misleading users with an inaccurate assessment of the models' abilities. 

\textbf{Results.}
Our evaluation results on the LHRS-Bench dataset are presented in Table \ref{tab:bench_result}, which clearly demonstrates that \MODELNAME~stands out as the best-performing approach overall. \MODELNAME~achieves the highest overall accuracy of 34.93\%, a significant margin above all other methods. It not only delivers top performance in areas like identity and reasoning but also maintains a balanced distribution of strong scores across multiple dimensions. Notably, it even outperforms the closed-source models, GPT-4o-mini and Claude-3, highlighting the importance of RS-specific training for interpreting RS images. Nearly all the MLLMs, except GPT-4o-mini, struggle to recognize the resolution of RS images, which is understandable given the lack of relevant datasets for training. As a RS chatbot, while \MODELNAME~demonstrates strong effectiveness in RS image understanding, these results also highlight promising directions for enhancing the general recognition capabilities of RS MLLMs, particularly in areas such as orientation, object counting, and other complex tasks.

\begin{table}
\centering
\caption{Comparison results on the LHRS-Bench dataset with different open-source and closed-source MLLMs (unit: \%).}
\label{tab:bench_result}
\resizebox{\textwidth}{!}{%
\begin{tabular}{@{}lcccccccccccc@{}}
\toprule \rowcolor{mygray}
Method             & Identity       	& Color     & Orientation  	& Shape  	 	& Area          	& Resolution   	& Modality       	& Location  	     & Distance       & Quantity     & Reasoning	 & OA           \\ \midrule
Qwen-VL-Chat          & 25.87 & 17.70 & \underline{10.26} & \textbf{45.95} & 17.33 & 0.00  & 0.00  & 16.67 & 9.09  & 8.76  & {43.48} & 24.93\\
LLaVA-1.5       & 29.34 & \underline{21.24} & \textbf{12.82} & 27.03 & 24.00 & 0.00  & 4.35  & 25.98 & \textbf{22.73} & \textbf{21.90} & \underline{58.70} & 28.26\\
InternLM-XComposer & 5.05  & 6.19  & 2.56  & 18.92 & 1.33  & 0.00  & 0.00  & 4.41  & 4.55  & 1.46  & 6.52  & 4.64\\
GeoChat         & 20.66 & 9.73  & 7.69  & 32.43 & 14.67 & 0.00  & 13.04 & 12.25 & 9.09  & \underline{9.49}  & 45.65 & 19.86\\        
gpt-4o-mini     & \underline{30.13} & 19.47 & 7.69  & \underline{43.24} & \textbf{29.33} & \textbf{28.57} & \underline{21.74} & \textbf{29.90} & 13.64 & 8.03  & 39.13 & \underline{30.00}\\
Claude-3-Opus   & 23.82 & 18.58 & 2.56  & \underline{43.24} & 18.67 & \underline{4.76}  & 17.39 & 26.47 & 4.55  & 6.57  & 30.43 & 23.19\\
\rowcolor{cyan!50}
% LHRS-Bot        & 0.32  & 0.00  & 0.00  & 0.00  & 0.00  & 0.00  & 0.00  & 0.98  & 0.00  & 0.00  & 0.00  & 0.29\\   
\MODELNAME     & \textbf{36.12} & \textbf{27.43} & 5.13  & 40.54 & \underline{25.33} & 0.00  & \textbf{26.09} & \underline{26.96} & \underline{18.18} & {8.76}  & \textbf{63.04}  & \textbf{34.93}\\ \bottomrule

\vspace{10mm}
\end{tabular}%
}
\end{table}

\subsection{Multimodal dialogue in RS understanding}
\label{subsec:Qualitative}

\begin{figure}[t]
  \centering
  \includegraphics[width=\linewidth]{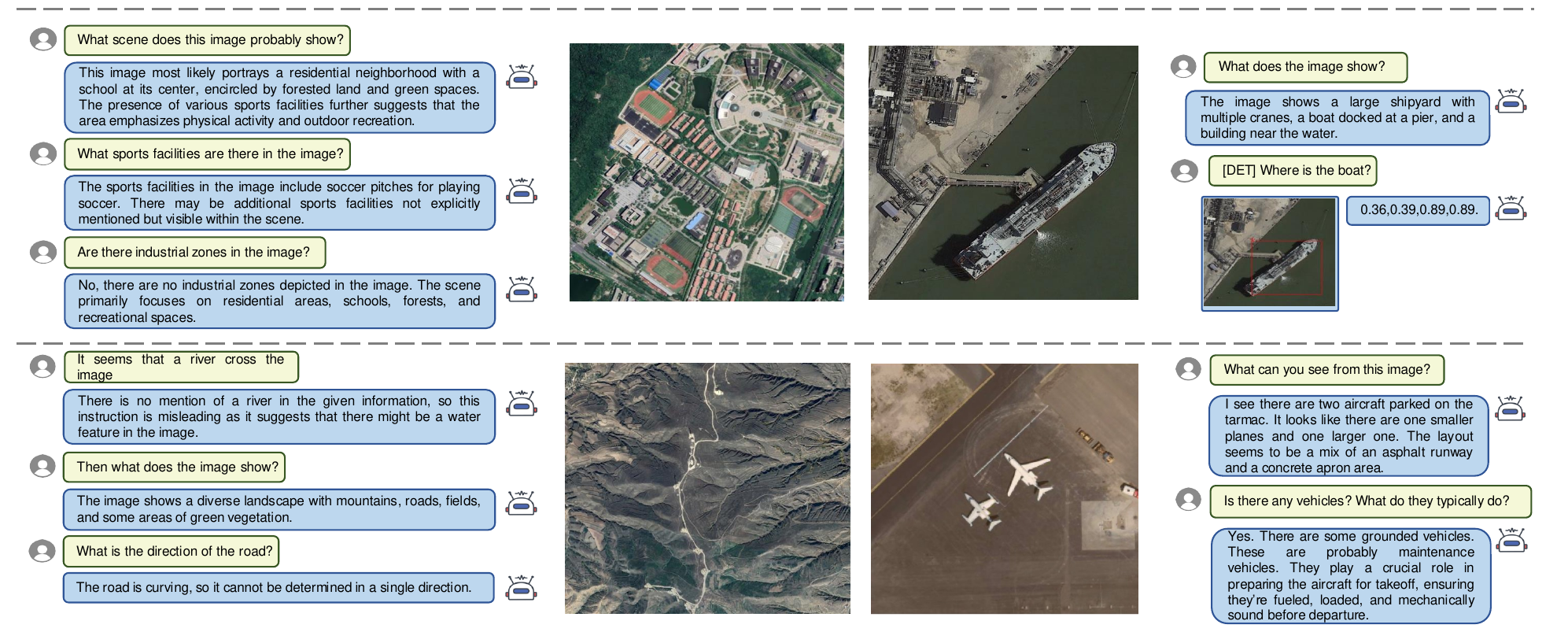}
  \caption{Four conversation examples between user and~\MODELNAME.}
  \label{fig: Conv. examples}
\end{figure}

We present conversation examples to qualitatively demonstrate~\MODELNAME~performance in RS image interpretation. As shown in Fig. \ref{fig: Conv. examples}, \MODELNAME~can clearly and thoroughly describe RS image scenes, showcasing its excellent recognition of RS features. It can engage in conversation with users, following instructions for tasks like object localization and reasoning. More importantly, it does not simply respond blindly to user queries. Thanks to the utilization of the more balanced instruction data with negative samples, it can assess the validity of user instructions and is capable of providing negative responses or refusing to answer, demonstrating high reliability. 
% \begin{wrapfigure}{r}{0.55\textwidth}
%     \begin{minipage}{0.55\textwidth}
%         \centering  
%         \vspace{-5mm}
%         \scalebox{1.00}
%         {
%             \includegraphics[width=\textwidth]{loss_curve.pdf}  
%         }
%         \captionof{figure}{The loss curve for different image-text pre-training datasets at stage 1.}
%         \label{fig:loss_curve}
%         \vspace{-25mm}
%     \end{minipage}
%     \vspace{15mm}
% \end{wrapfigure}or refusing to answer, demonstrating high reliability. 

\subsection{Ablation analysis}
\label{subsec:Ablation}

\textbf{Effectiveness of improved caption quality.} 
To assess the effectiveness of data recaptioning, we compare the results from models trained on both the original
and recaptioned datasets, as shown in Table \ref{tab:dataset ablation}. 
Each dataset is used separately for pretraining, while all other experimental conditions remain consistent.
For different tasks, we calculate the mean accuracy across all datasets outlined in Section \ref{subsec:setup}. 

\begin{figure}[H]
    \centering  
    \vspace{-5mm}
    \scalebox{0.6}
    {
        \includegraphics[width=\textwidth]{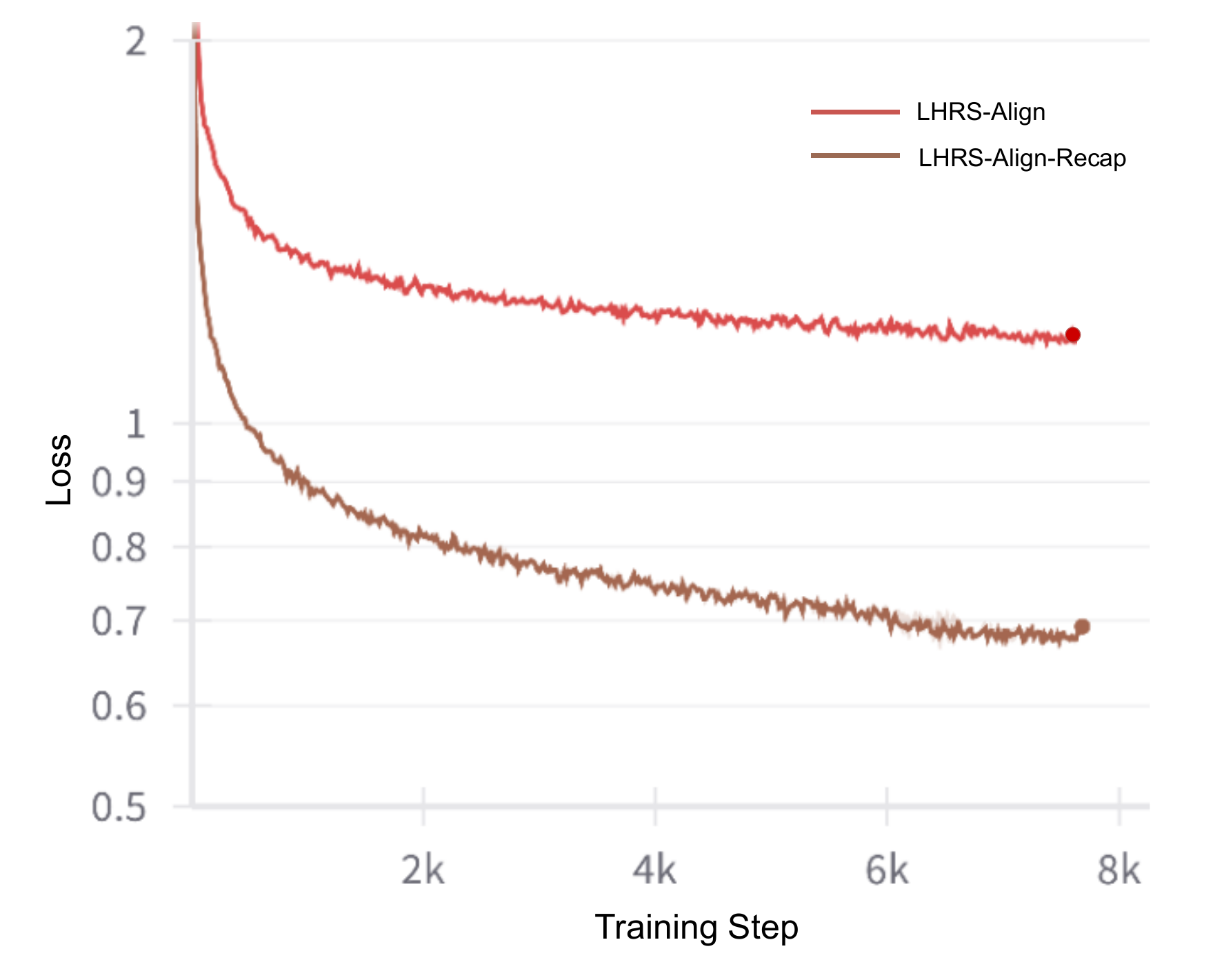}  
    }
    \captionof{figure}{The loss curve for different image-text pre-training datasets at stage 1 training.}
    \label{fig:loss_curve}
    % \vspace{-25mm}
\end{figure}
\vspace{5mm}

The training curve in Fig. \ref{fig:loss_curve} clearly shows that the model trained on LHRS-Align-Recap exhibits a lower loss during the first stage of training, indicating better vision-language alignment. Ablation results in Table \ref{tab:dataset ablation} confirm that model performance consistently improves across all tasks when pre-trained with the proposed LHRS-Align-Recap dataset, highlighting the effectiveness and importance of generating better captions. The most significant improvement is observed in classification accuracy, with a 20.39\% increase, demonstrating a substantial enhancement in RS scene recognition capability. Notably, the model trained with LHRS-Align-Recap achieves a 9.1\% accuracy improvement in the visual grounding task. 
This improvement demonstrates that, compared to using an LLM for caption generation, leveraging an MLLM that can actually see images allows for more accurate capture of spatial information about geographical objects.

\begin{table}[H]
\centering
\caption{Accuracies (\%) across various tasks for models trained with different pretraining datasets.}
\label{tab:dataset ablation}
\centering
\scalebox{0.88}{
\begin{tabular}{@{}lcccccc@{}}
\toprule \rowcolor{mygray}
Dataset  & Classification & VQA   & Visual Grounding   & LHRS-Bench\\ \midrule

LHRS-Align       & 56.21     & 90.30     & 78.26     & 21.02 \\
% LHRS-Instruct?  & ?     & ?     & ?     & ? \\
\rowcolor{cyan!50}
LHRS-Align-Recap    & 76.60     & 90.59     & 87.36     & 34.93 \\ \bottomrule

\end{tabular}%
}
\vspace{-5mm}
\end{table}

\textbf{Effectiveness of model architecture.}
To verify the effectiveness of the proposed MoE Vision Perceiver, we train the model with the vanilla vision perceiver and the MoE vision perceiver using the same data and hyperparameter settings, and then evaluated them on each downstream task. 
The results, shown in Table~\ref{tab:model ablation}, indicate that the MoE Vision Perceiver consistently outperforms the vanilla vision perceiver across all tasks, with a margin of up to 6.24\% in visual grounding and 4.47\% in LHRS-Bench. This suggests that the MoE architecture has indeed expanded the model's memory capacity, enabling LHRS-Bot-Nova to focus on more visual details.

\begin{table}[H]
\centering
\caption{Accuracies (\%) across various tasks for both the Vanilla Vision Perceiver and the proposed MoE Vision Perceiver.}
\label{tab:model ablation}
\centering
\scalebox{0.88}{
\begin{tabular}{@{}lcccccc@{}}
\toprule \rowcolor{mygray}
Method   & Classification & VQA   & Visual Grounding   & LHRS-Bench\\ \midrule

Vanilla Vision Perceiver & 74.19 & 90.44     & 81.12     & 30.46 \\
\rowcolor{cyan!50}
MoE Vision Perceiver    & 76.60     & 90.59     & 87.36     & 34.93 \\ \bottomrule

\end{tabular}%
}
\vspace{-5mm}
\end{table}

% \section{Discussion}
% \label{sec6}

\section{Conclusion}
\label{sec:conclusion}
We propose \MODELNAME, which integrates visual signals with language expression to achieve unified RS image interpretation and understanding under human instructions. With higher information density and quality in the synthetic image-caption alignment data, the improved instruction data targeted at spatial reasoning, and the vision-centric architecture design, LHRS-Bot-Nova demonstrates strong performance not only in scene classification, visual grounding, and question-answering tasks but also outperforms other MLLMs in general RS interpretation. 
Additionally, the results of our systematic evaluation across different models and tasks provide a reliable reference for future model selection and enhancement.

Despite the impressive performance of \MODELNAME, it shares limitations common to LLMs, such as susceptibility to hallucinations. We believe that designing a more rigorous alignment data curation pipeline, along with improved training strategies such as preference alignment, can further enhance the performance of MLLMs in interpreting RS images.
% future work: unbalance of positive and negative samples in SFT.
% limits: hallucination in LHRS-Alignv2, too much length
% larger amount of training data with high-quality images. 
% Instruction data low accuracy and hallucination
% Multiple-choice limits

% \section*{Data availability}

\section*{Acknowledgments}
This study was supported by the National Natural Science Foundation of China
(Grant No. 42071297, 42471410) and the AI and AI for Science Project of Nanjing University (Grant No. 02091480605203).

\clearpage

% \appendix
% \section{}
% \label{}

\clearpage
\bibliographystyle{elsarticle-harv} 
\bibliography{references}

\end{document}